# Polyth-Net: Classification of Polythene Bags for Garbage Segregation Using Deep Learning

Divyansh Singh
*Computer Science Engineering*
*The LNMIIT*
Jaipur, India
18ucs127@lnmiit.ac.in

*Abstract*— Polythene has always been a threat to the environment since its invention. It is non-biodegradable and very difficult to recycle. Even after many awareness campaigns and practices, Separation of polythene bags from waste has been a challenge for human civilization. The primary method of segregation deployed is manual handpicking, which causes a dangerous health hazards to the workers and is also highly inefficient due to human errors. In this paper I have designed and researched on image-based classification of polythene bags using a deep-learning model and its efficiency. This paper focuses on the architecture and statistical analysis of its performance on the data set as well as problems experienced in the classification. It also suggests a modified loss function to specifically detect polythene irrespective of its individual features. It aims to help the current environment protection endeavours and save countless lives lost to the hazards caused by current methods.

## I. INTRODUCTION

Cities and urban population have been combating the consequences of using polythene bags since they are been mass produced for public use. Polythene, an acronym for Polyethylene is a hydrocarbon and is generally a waste product resulting from Domestic waste and seldom from Industrial waste. They are very difficult to dispose of and take very long time to decompose naturally. Burning them causes toxic gases to be formed and hence is not a solution. They are also toxic to water bodies and his almost impossible to get rid of.

Annually approximately 500 billion polythene bags are used worldwide. More than one million bags are used every minute. A plastic bag has an average "working life" of 15 minutes.

Polythene is recyclable and recycling is the best practice to slow down and control its aftermaths. But it is only feasible and efficient if polythene bags are segregated from other waste materials effectively.

The primary method currently used for segregation of polythene bags is manual handpicking of the bags. This method is inefficient due to large amounts of garbage as well as it proves to be major health hazards for the human workers since it involves direct contact with garbage which includes all sorts of waste materials including industrial toxic wastes and disease-causing microbes. These hazards are responsible for many health-related ailments in the lives of these workers due to poor sanitization and hygiene in their job. This method is also comparatively expensive as it includes wages for all such workers. Certain chemical-based methods are also suggested but they also prove to be costly and impractical for industrial deployment. So, I worked on developing an Artificial Intelligence based model which can recognize polythene bags from their images with minimal human contact for their segregation and hence be used as a hygienic substitute instead of handpicking as it only involves machine and less labour.

The reason for using Artificial Intelligence is simple as it minimizes human effort and is also less expensive than the other methods in current use.

The main objective is to develop a model which is able to perform image-based classification of polythene bags from other components efficiently. It also aims to reduce human effort less costly than the current practices. I also must focus on efficiency of the model on real world data and evaluate the model's performance and reduce the error as much as possible within the scope of my research. This also includes selection of an optimal loss function to achieve the better performance.

## II. DEEP LEARNING

Deep Learning is a field of artificial intelligence which focuses on classification or regression of data. It is a broader subset of machine methods based on artificial neural networks with representation learning. Learning can be supervised, semi-supervised or unsupervised. The problem of identifying polythene bags is a supervised learning problem and hence the model is trained on data containing images of all the classes and their corresponding labels. A Deep Learning model consists of weights stored in neural nets which is used to compute the result of the classification/regression task. The model uses mathematical functions and calculus to propagate and image through the network and then compute the partial derivatives for each parameter using backpropagation[1] and then update the parameters to minimize the cost function or the loss function. This is repeated several times until an optimum minimum is found for the loss function and hence the network is able to predict new data using these learnt parameters. The neural network used in this problem is a Convolutional Neural Network which specifically designed for image-based classification tasks.

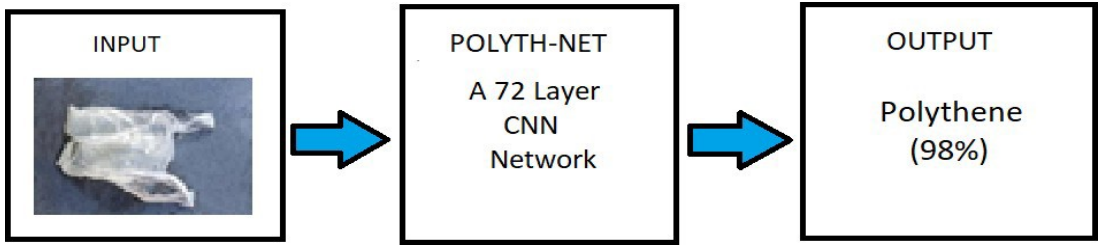

Fig. 1. Flowchart depicting the Working of the model

A deep learning model has been deployed in this problem as it proves to be the most accurate machine learning algorithm in the image-classification based problems.

*A. Problem Definition*

The polythene detection is categorical classification problem where input X is an RGB image array of 3 channels (Red, Green, Blue) and outputs y $\in$ {0,1,2} where the classes {0,1,2} are labelled as: 0- non-plastic, 1-other types of plastic excluding polythene and 2- polythene. The task of the model is to process the image through the network over training with multiple epochs and minimize the loss function.

The model will output the results in a one-hot vector representing the probabilities of each of the classes being present in the image. The highest probability will be selected as the final output; however, a threshold can be applied to the predictions to prevent misclassification of polythene bags class.

*B. Loss Function*

Loss function is the function computed for estimation of the deviation of the model from an 'ideal' model. The cost function reduces all the various good and bad aspects of a possibly complex system down to a single number, a scalar value, which allows candidate solutions to be ranked and compared.

Categorical cross entropy[2] is suitable for multiclass classification. This is achieved by the loss function:

$$L(y, \hat{y}) = -\sum_{j=0}^{M}\sum_{i=0}^{N}(y_{ij} * log(\hat{y}_{ij}))$$

Where y, ŷ are true and predicted values respectively and is the batch size and N is number of classes respectively.

But as this model is specifically made for detecting polythene bags, hence the Loss Function has been modified to be harsher when a polythene bag is classified incorrectly by giving additional weight to the loss of polythene class i.e. 2. This is achieved by the loss function:

$$-\sum_{j=0}^{M}[\sum_{i=0}^{1} y_{ji} * log(\hat{y}_{ji}) + \lambda * y_{j2} * log(\hat{y}_{j2})]$$

Where predictions and labels are in the form of matrices of order (Batch size x no. of classes), non-polythene classes are labelled as 0,1 and Polythene is labelled as 2.

The additional weight is controlled by the parameter $\lambda\lambda$ which controls the weight of the loss to the Polythene class, values of $\lambda\lambda$ greater than 1 penalizes the model more for inaccurate prediction of the polythene class and values less than 1 neglects or reduces the corresponding loss. Very large values of $\lambda\lambda$ is strongly discouraged as the model will be unable to accurately predict other classes.

The value of $\lambda\lambda$ was kept as 1.25 during training of the model which made the model more sensitive to loss in the Polythene Class but also not completely neglecting the loss of other two classes and hence performance of the model is expected to increase specifically in recognizing polythene bags and its features.

*C. Model Architecture and Training*

Polyth-Net is a 72 layered Xception [3] model trained on a dataset containing the 3 classes to be classified. Xception [3] model uses depth wise separable convolutions to perform better on image classification tasks than standard convolutional neural networks. See Table 1 for reference.

| INPUT Layer |
| --- |
| XCEPTION CNN |
| FC Dense Layer 1,0.25 Dropout |
| FC Dense Layer 2,0.25 Dropout |
| SOFTMAX LAYER (OUTPUT) |

Table 1: Architecture of the model

The model's neural architecture has a Xception[3] model without the original top layers followed by 2 Fully Connected Dense layers with ReLu[4] as an activation function and then the final SoftMax[4] Output layer. It has total 21.3 million parameters out of which about 21.2 million are trainable. Regularization in the form of Random Dropout[5] with a keep probability of .25 has been implemented on both the Dense layers to prevent overfitting on the training data. The model was built as a sequential CNN model and the loss function to be minimized during training and validation of the model

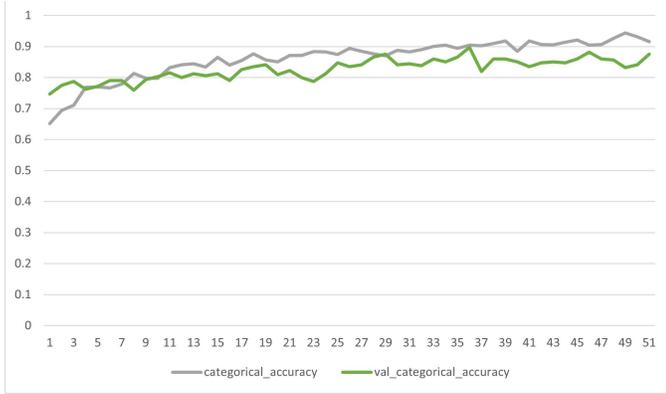

Fig. 2. Graph of accuracy vs epoch while training of the model

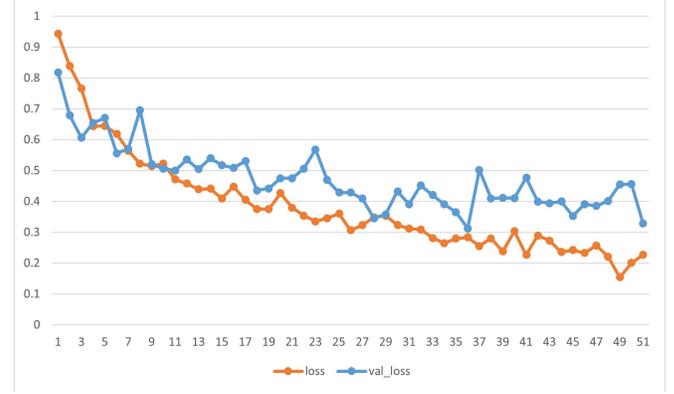

Fig. 3. Graph showing loss vs epoch while training.

was used as described above with the value of $\lambda$ to be kept as 1.25.

Also, Additional Regularization in the form of Early Stopping[6] on the validation set was used to further prevent overfitting on the training set.
The weights for the neural network were initialized using Xavier Initialization for random initialization of weights[7] which makes sure that each activation neuron is independent of the initial weights of other neurons.

The model is trained end to end using a Adam[8] optimizer with the standard parameters($\beta_1 = 0.9$ and $\beta_2 = 0.999$). The training was done with a batch size of 32 and the initial learning rate was kept as 0.001 which was reduced by a factor of 10 after every epoch of training.

The model was trained 5 times to achieve similar results to prevent the training to assume a saddle point as a minima.
A saddle point is a point of local minima which prevents the model from training and progressively train further.
Training multiple times reduces the chances of model getting stuck at saddle point or local minima in the feature dimensional space.

### III. DATA

*A. Preprocessing:*

The image is resized to (224,224,3) RGB image by using Linear Interpolation technique and then preprocessed further. The images were scaled by a value of 255 to ensure that pixel values are in range (0,1) and then normalize the images by each channel with the standard values from ImageNet[9] dataset. The model was tested by converting the images to grayscale format, but it did not produce significantly accurate results and hence the RGB format was preferred.

*B. Training*

The dataset was built using images collected under my scope of research from random items found in garbage heaps. The data was augmented randomly to increase the efficiency of the model and prevent overfitting on the training data. There were about 400 images of the polythene class, 1200 images of the non-plastic class and 500 of the other plastic classes. The model was trained using augmented images from the dataset which were flipped horizontally, randomly rotated within a range of (0,180) and randomly zoomed in the range (0.4,1.4) The model was trained with a batch size of 32 and 25 training steps per epoch. The metrics chosen for the training were accuracy and f1 score of the model's predictions. See Fig. 2 and Fig. 3 for reference.

The validation data was a subset of the dataset having about 577 images and were randomly augmented with same parameters as the training dataset.

There exist many non-plastic bags which may look like a polythene bag in their shape and size. To ensure that the model learns not to misclassify them as polythene ones, the training data contains significant number of these paper/cloth bags samples and their augmentations. This makes the model more robust to such common mistakes.

| Dataset | Accuracy | F1 Score |
|---------|----------|----------|
| Training | 96.39 | 0.9645 |
| Testing | 88.84 | 0.8977 |
| Validation | 95.52 | 96.33 |

Table 2. Performance evaluation of the dataset

*C. Testing*

The test dataset was consisting about 50 images from each class and was pre-processed same as the training dataset and augmented randomly as the training dataset with the same method. The testing dataset was developed by collection of images sampled randomly from garbage items and labelling them correspondingly and was collected separately from the training and validation data.

## IV. PERFORMANCE AND LIMITATIONS

Polyth-net performed reasonably good on the test dataset in comparison of the training statistics, overfitting is not very prominent as the model is performing good on new data. The model achieved a f1-score of 0.90 on test data and an accuracy if 88.9% which is slightly less than the validation data and shows a slight case of overfitting but not very high and to be considered a problem. The regularization techniques implemented on the model has shown significant results in minimizing the difference between training and testing statistics. The model also generalizes well between plastic and paper bags which is also a good sign of performance

The model has certain limitations as it has been trained on a medium size dataset, while efforts have been made to minimize overfitting on the dataset but still it may exist in some forms. Also. the model has large numbers of parameters and is expensive to compute even though it has less layers than most industrial grade classification models.

The model has also been found to misclassify some glass-items which are deformed and may look like a polythene bag from its visual analysis. Some Logistical issues may occur in the implementation like the time taken to compute the output could be unfeasible, but it can be solved by a powerful computer. Models with lower number of parameters can be used as a tradeoff for the accuracy.

## V. SUMMARY AND CONCLUSION

Polyth-Net, A convolutional network presented in this paper can be implemented to be used in real-time segregation tasks of polythene bags by some additional modification as the performance of the model is satisfactory on test-data. The loss function is customizable and is suitable for more specific identification of polythene bags to ensure a more robust performance on test data. This also includes other non-plastic bags like cloth and paper bags. The performance of Polyth-Net can be greatly improved by training on a much larger dataset comprising all kinds of images which may exist in practice and can reduce overfitting also make the model more robust to new samples and have a higher test accuracy, hence more efficient. Also, the model can only detect single objects which can also be a problem in its implementation, but this problem can be removed by adequate pipelining methods in practice. The deployment of this model will be able to solve the current situation which is affecting most of the areas in the world. Hygienic issues would be resolved. See Table 2 for test scores.